# Traceable Fault Diagnosis for Battery Energy Storage Systems via Retrieval-Augmented Multi-Agent O&M Assistant

Jiangdi Ru[1], Bing Li[2], Yage Huang[2], Ding Wang[1,*], Keru Hua[1]

## ABSTRACT

Large-scale battery energy storage systems (BESSs) require O&M decisions that combine alarms, cell-level measurements, device topology, diagnostic tables, historical cases, and maintenance documents. Monitoring platforms can flag threshold violations, but they often cannot explain whether voltage inconsistency, resistance drift, short-circuit risk, capacity divergence, or thermal abnormality needs intervention. This digest presents a traceable BESS fault-diagnosis assistant that uses retrieval-augmented multi-agent reasoning to connect operational data, domain knowledge, visual evidence, and report generation. Reliability is improved through BESS-specific task routing, schema-constrained natural-language database access, hybrid text-image retrieval, and evidence-based answer synthesis. Preliminary internal evaluation is reported for routing, database access, and diagnostic reasoning.

## INTRODUCTION

Battery energy storage systems are increasingly deployed for renewable integration, peak shaving, frequency regulation, and grid flexibility. Their availability depends on coordinated operation of cells, modules, racks, BMS, PCS, EMS, thermal control, fire protection, and O&M platforms. As plant scale increases, early faults may appear as weak but persistent patterns: cell-voltage inconsistency, resistance growth, temperature rise, static-voltage reversal, recurrent alarms, or capacity divergence.

Current O&M workflows remain fragmented. Operators inspect alarms in one platform, retrieve curves from a time-series database, search manuals or standards elsewhere, and consult experts for final judgment. Data-driven battery models can detect anomalies or estimate SOH, but most output a score or label rather than a traceable maintenance decision. General LLMs are also unsafe without private knowledge grounding, controlled database access, conservative risk reasoning, and auditable evidence.

[1] Corporate Research Center, Midea Group, Foshan528311, People's Republic of China
[2] New Energy Business Division, Midea Group, Foshan528311, People's Republic of China
[*] Corresponding author: ding2.wang@midea.com

This work bridges algorithmic battery diagnosis and practical O&M decision support. The assistant is designed around BESS-specific routes, data tables, fault categories, and evidence bundles. The contributions are: (i) a BESS O&M task model with complexity-aware routing; (ii) schema-constrained database access; (iii) hybrid text-image RAG for manuals, procedures, and failure-analysis figures; and (iv) multi-agent evidence synthesis for traceable conclusions and actions.

## RELATED WORK

Battery safety and fault diagnosis have been studied using model-based, statistical, and data-driven methods. Small-fault detection can use electrochemical models and residual tests [1], while large-pack internal short-circuit detection may require additional force or gas sensing [2]. Field-data studies highlight online health monitoring and weakest-cell behavior in real systems [3]. For grid BESS O&M, inconsistency-driven methods show that electrical, thermal, and aging indicators must be translated into explainable maintenance actions rather than numerical alarms [4].

RAG and agent methods provide the interface layer needed for such workflows. RAG grounds generation in documents [5], dense retrieval improves semantic recall [6], and BM25 remains useful for exact device identifiers and fault terms [7]. Multimodal RAG supports figures in maintenance procedures and failure-analysis reports [8]. ReAct, Toolformer, and Tree-of-Thoughts connect reasoning with external actions [9]-[11], while Text-to-SQL research shows the need for schema linking and constrained generation [12], [13]. BatteryAgent combines physical features, explainable ML, and LLM reasoning for battery diagnosis [14]. Our focus differs by integrating database query, knowledge retrieval, visualization, and multi-agent evidence synthesis for BESS O&M.

## BESS O&M TASK MODEL AND SYSTEM ARCHITECTURE

The assistant organizes user requests into three BESS business routes: alerting, troubleshooting, and station/device time-series analysis. These routes map to alarm-event tables, diagnostic-result tables, and equipment measurement tables, keeping both the system and the paper anchored in energy-storage operations.

Table I: BESS O&M tasks supported by the assistant.

| **Task category** | **Typical input** | **Data/evidence source** | **Expected output** |
|---|---|---|---|
| Alarm analysis | Station/rack/pack, time window | alarm_event and abnormality tables | Severity, recurrence, risk note |

| Task category | Typical input | Data/evidence source | Expected output |
|---|---|---|---|
| Time-series trend | Device ID, signal, period | box/cluster/BMU measurements | Trend, chart, abnormal interval |
| Troubleshooting | Observed symptom or weak signal | diagnostic tables, RAG evidence | Candidate causes and actions |
| Procedure/report | Fault term or multi-source findings | manuals, cases, images, DB rows | Grounded explanation or report |

Fig. 1 summarizes the workflow. A complexity-aware router sends simple tasks to a single-agent supervisor for DATABASE, RETRIEVE, CHART, or DIRECT actions, and sends complex diagnostic questions to a deep-research path with pre-retrieval, draft generation, supervisor-researcher coordination, finding compression, and final report synthesis. Intermediate states are logged for audit and expert review.

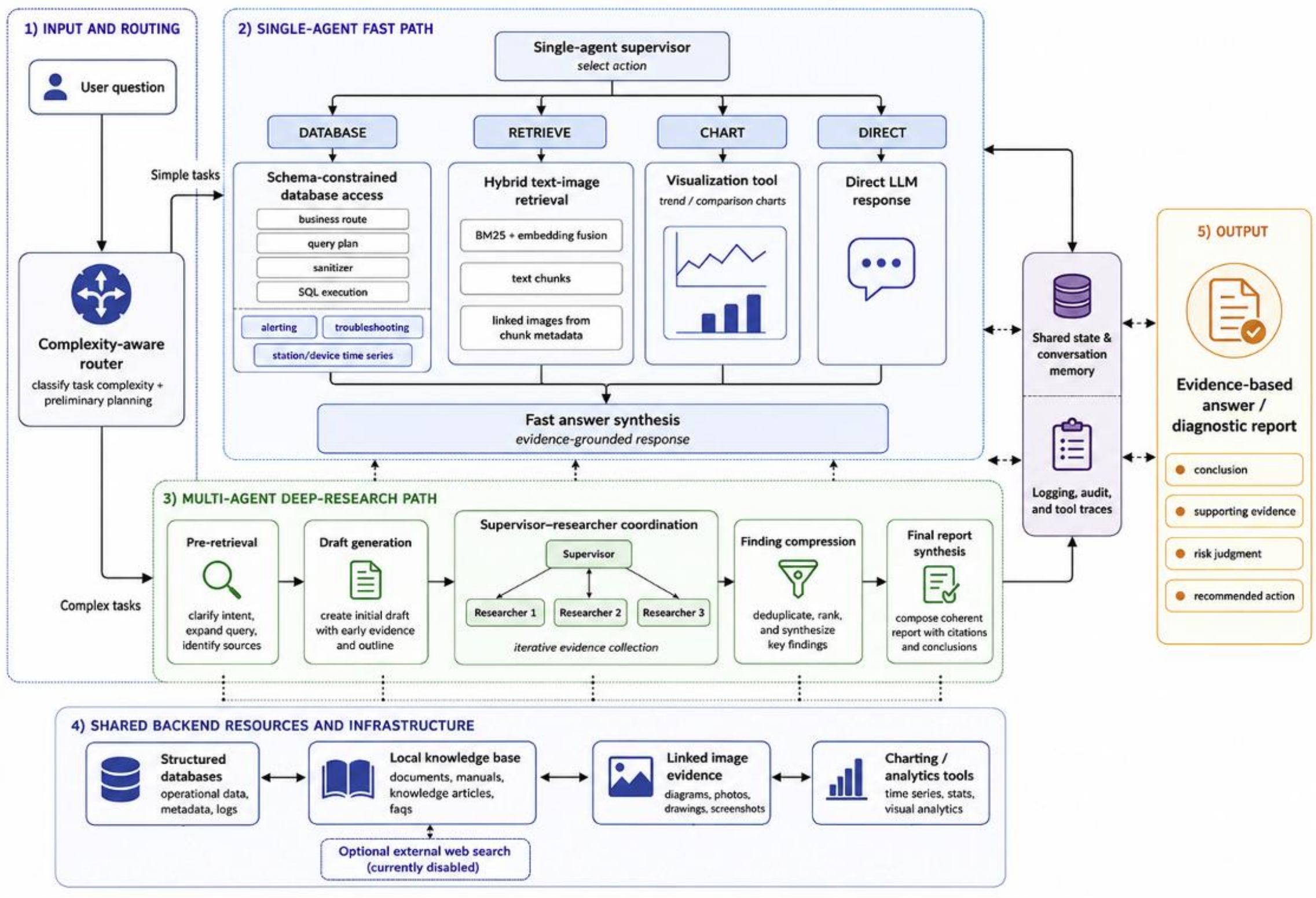

Figure 1: Overall BESS intelligent O&M assistant architecture.

## METHODOLOGY

The database module follows a plan-then-execute design. The LLM predicts the business route and a structured query plan containing table, fields, filters, time range, ordering, and limit. A deterministic sanitizer checks schema allowlists and route-table mappings, removes illegal fields, normalizes aliases, clamps unsafe limits, and builds executable SQL. Returned rows are converted into an evidence bundle before answer generation.

The retrieval module uses a hybrid text-image pipeline. Offline, manuals, standards, procedures, and cases are parsed into text chunks with linked image metadata. Online, dense retrieval and BM25 retrieval are fused as s(d) = alpha BM25(d) + (1-alpha) Emb(d) after normalization. Text grounds the answer, while linked images provide visual evidence for procedures, morphology analysis, and waveform interpretation.

The multi-agent path is reserved for high-complexity questions such as cross-document root-cause analysis or symptoms without a clear device ID. A supervisor decomposes the request, dispatches researcher agents, compresses findings, and generates a diagnostic report containing conclusion, evidence, risk judgment, and recommended action.

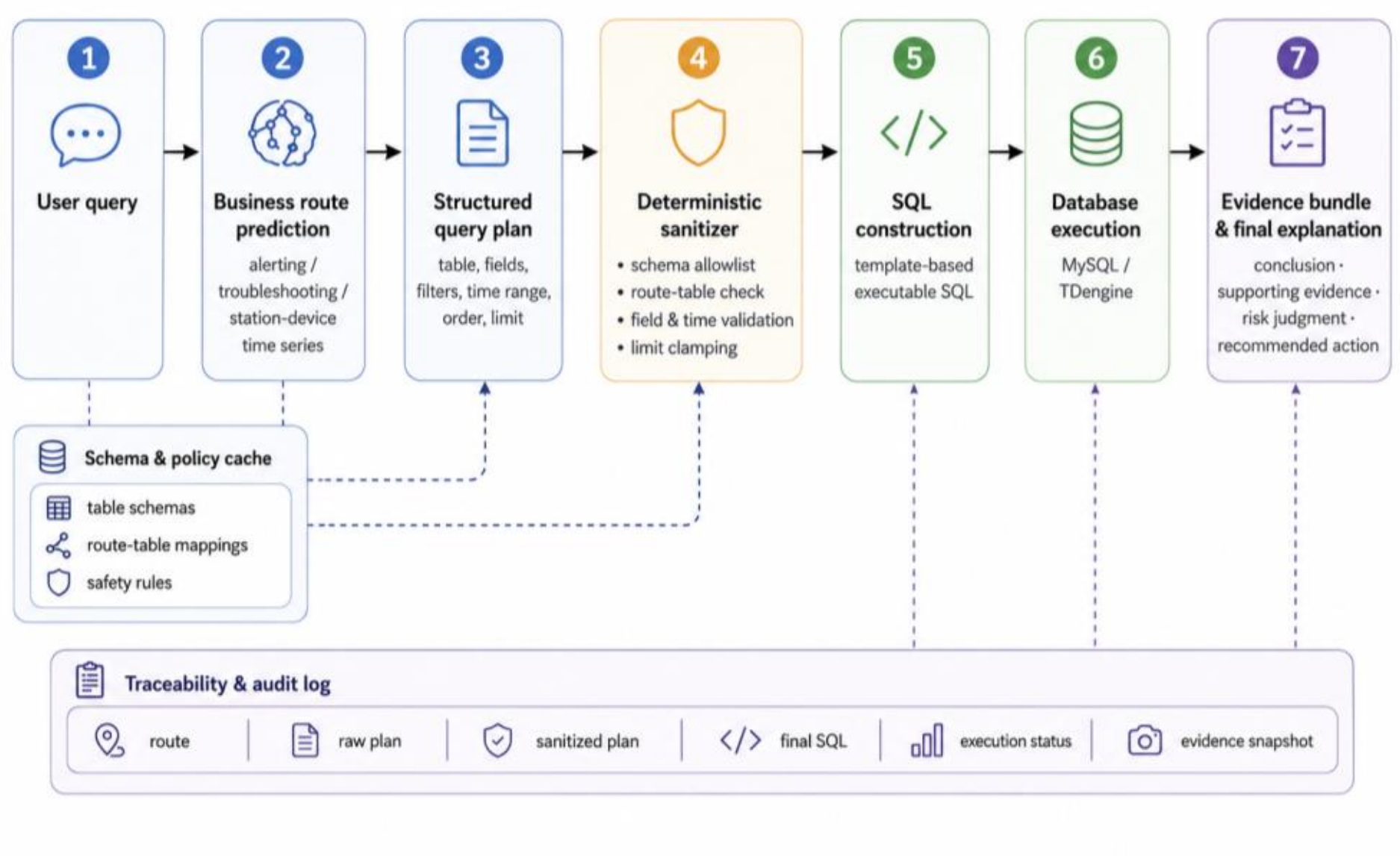

Figure 2: Schema-constrained database query and evidence-fusion pipeline.

## PRELIMINARY EVALUATION AND CASE STUDIES

A preliminary internal evaluation used anonymized operational data and a private BESS maintenance knowledge base. The resource pool contains three business routes, seven queryable tables, 99 documents, 6,741 text chunks, 717 images, and 486 image-linked chunks. The task suite covers routing, database-query, and diagnostic-reasoning questions; final station scale, sampling period, expert-label protocol, and screenshots will be added after clearance.

For database access, SQL-ready plan success denotes the fraction of query plans that pass route/schema checks and can be deterministically converted into executable SQL without manual repair. The schema-constrained module achieves 100% success by normalizing aliases and checking table-field mappings, while removing schema validation yields 0% safe SQL-ready success because raw LLM plans often contain invalid fields, route-table mismatches, or unsupported query scopes.

Table II: Preliminary quantitative comparison with ablated variants.

| Component | Full system | Ablated variant | Main observation |
|---|---|---|---|
| Routing | 70.0% action accuracy; quality 4.44; 40.01s | w/o routing: 20.0%; quality 3.20; 87.45s | Better tool selection and lower latency. |
| Database access | 100% Safe query-plan success; 43.99s | w/o schema validation: 0%; 32.39s | Only validated query plans are counted. |
| Diagnosis | Quality 4.80; 95.33s | w/o multi-agent: quality 3.60; 60.99s | Improves completeness and traceability. |

Beyond the quantitative comparison, Fig. 3 presents four representative BESS O&M cases generated by the current prototype. The cases cover ISC-risk evidence retrieval based on micro-short-circuit scores, cluster-level telemetry query and trend visualization, BESS knowledge-base retrieval with document and image evidence, and recurrent alarm and abnormality analysis for a battery pack. These cases show that the assistant can transform natural-language O&M questions into schema-constrained database queries, chart generation, knowledge retrieval, and structured diagnostic summaries. Sensitive station names, device identifiers, and numerical values are anonymized or partially masked for confidentiality.

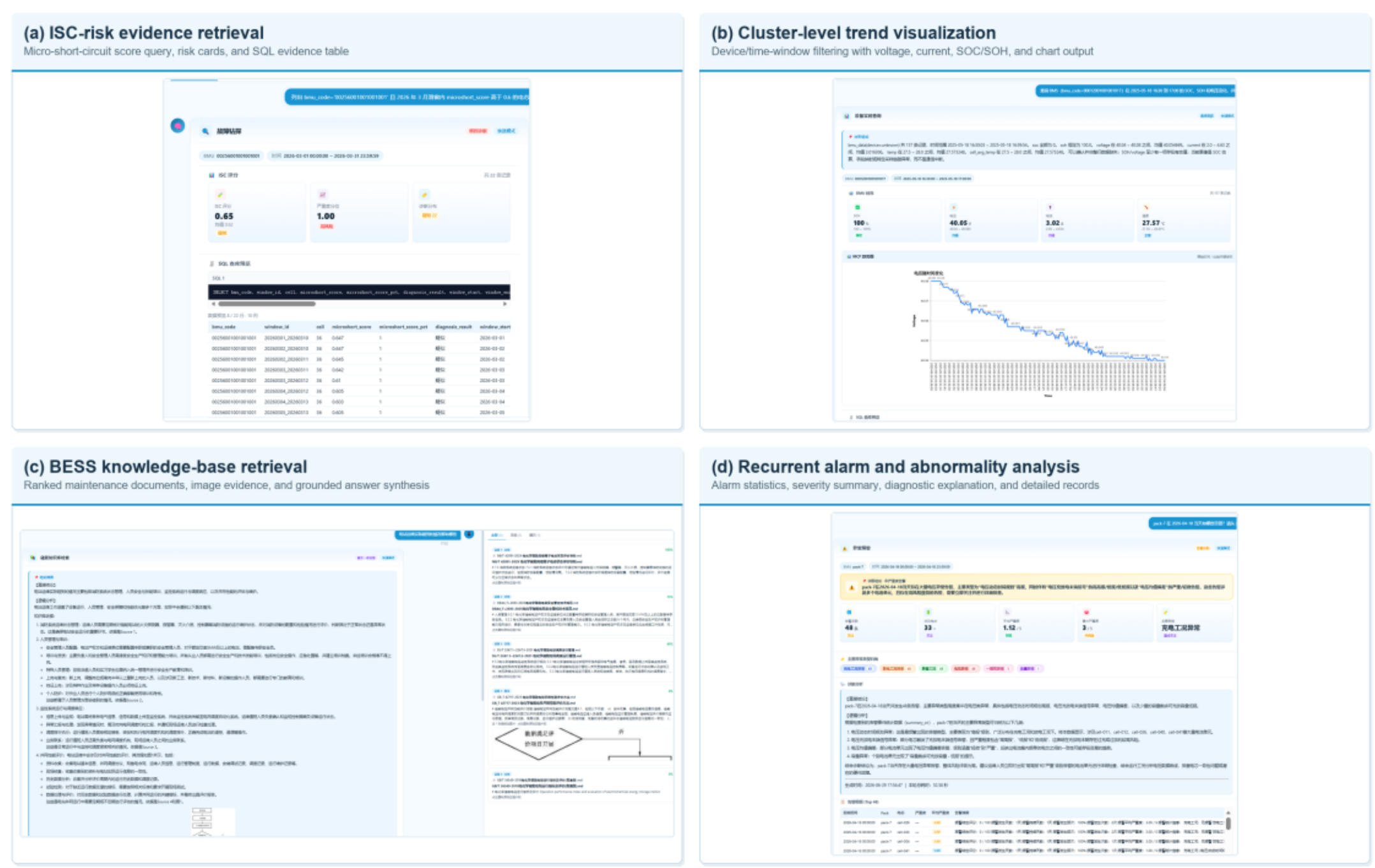

Figure 3: Representative BESS diagnostic and O&M cases generated by the prototype.

## CONCLUSIONS AND FUTURE WORK

This digest presents a BESS-oriented intelligent O&M assistant for traceable battery fault diagnosis. It keeps energy-storage safety, operational data, fault categories, and maintenance actions at the center, while using LLM agents, RAG, and tools as supporting mechanisms. Detailed field statistics and additional expert-labeled case studies will be reported in the final paper..

**REFERENCES**

[1] L. D. Couto, J. M. Reniers, D. A. Howey, and M. Kinnaert, "Detection and Isolation of Small Faults in Lithium-Ion Batteries via the Asymptotic Local Approach," arXiv:2103.09936, 2021.

[2] T. Cai, P. Mohtat, A. G. Stefanopoulou, and J. B. Siegel, "Li-ion Battery Fault Detection in Large Packs Using Force and Gas Sensors," arXiv:2010.13519, 2020.

[3] J. Schaeffer et al., "Gaussian Process-based Online Health Monitoring and Fault Analysis of Lithium-Ion Battery Systems from Field Data," arXiv:2406.19015, 2024.

[4] J. Qu, Y. Wang, Y. Fu, P. Zhang, W. Li, and M. Li, "From Inconsistency to Decision: Explainable Operation and Maintenance of Battery Energy Storage Systems," arXiv:2601.03007, 2026.

[5] P. Lewis et al., "Retrieval-Augmented Generation for Knowledge-Intensive NLP Tasks," in Proc. Advances in Neural Information Processing Systems (NeurIPS), 2020.

[6] V. Karpukhin et al., "Dense Passage Retrieval for Open-Domain Question Answering," in Proc. EMNLP, 2020.

[7] S. Robertson and H. Zaragoza, "The Probabilistic Relevance Framework: BM25 and Beyond," Foundations and Trends in Information Retrieval, vol. 3, no. 4, pp. 333-389, 2009.

[8] W. Chen, H. Hu, X. Chen, P. Verga, and W. W. Cohen, "MuRAG: Multimodal Retrieval-Augmented Generator for Open Question Answering over Images and Text," arXiv:2210.02928, 2022.

[9] S. Yao et al., "ReAct: Synergizing Reasoning and Acting in Language Models," arXiv:2210.03629, 2022.

[10] T. Schick et al., "Toolformer: Language Models Can Teach Themselves to Use Tools," arXiv:2302.04761, 2023.

[11] S. Yao et al., "Tree of Thoughts: Deliberate Problem Solving with Large Language Models," arXiv:2305.10601, 2023.

[12] M. Pourreza and D. Rafiei, "DIN-SQL: Decomposed In-Context Learning of Text-to-SQL with Self-Correction," arXiv:2304.11015, 2023.

[13] H. Li, J. Zhang, C. Li, and H. Chen, "RESDSQL: Decoupling Schema Linking and Skeleton Parsing for Text-to-SQL," arXiv:2302.05965, 2023.

[14] S. Zhou, R. Liu, B. Su, J. Wang, Y. Wang, and B. Jiang, "BatteryAgent: Synergizing Physics-Informed Interpretation with LLM Reasoning for Intelligent Battery Fault Diagnosis," arXiv:2512.24686, 2025.

[15] C. Zhang, H. Zhao, and W. Zhang, "Accuracy and Robust Early Detection of Short-Circuit Faults in Single-Cell Lithium Battery," arXiv:2412.17234, 2024.

[16] C. Wong et al., "Health Feature Extraction from Battery Energy Storage System Field Fault Data," arXiv:2606.26347, 2026.